\documentclass[10pt,twocolumn,letterpaper]{article}

\usepackage[review]{}      

\usepackage{graphicx}
\usepackage{amsmath}
\usepackage{amssymb}
\usepackage{booktabs}

\usepackage{times}
\usepackage{epsfig}
\usepackage{float}
\usepackage{xfrac}
\usepackage{makecell}
\usepackage{multicol}
\usepackage{multirow}
\usepackage{verbatim}
\usepackage{ulem}
\usepackage{bm}

\usepackage[pagebackref,breaklinks,colorlinks]{hyperref}

\usepackage[capitalize]{cleveref}
\crefname{section}{Sec.}{Secs.}
\Crefname{section}{Section}{Sections}
\Crefname{table}{Table}{Tables}
\crefname{table}{Tab.}{Tabs.}

\def\cvprPaperID{*****} 
\def\confName{}
\def\confYear{}

\begin{document}

\title{$\rm D^2SL$: Decouple Defogging  and  Semantic Learning for Foggy Domain-Adaptive  Segmentation}

\author{Xuan Sun, Zhanfu An, and Yuyu Liu\\
BOE Technology Group Co., LTD.\\
{sunxuan@boe.com.cn}
}
\maketitle

\begin{abstract}
We investigated domain adaptive semantic segmentation in foggy weather scenarios, which aims to enhance the utilization of unlabeled foggy data and improve the model's adaptability to foggy conditions. 
Current methods rely on clear images as references, jointly learning defogging and segmentation for foggy images. Despite making some progress, there are still two main drawbacks: (1) the coupling of segmentation and defogging feature representations, resulting in a decrease in semantic representation capability, and (2) the failure to leverage real fog priors in unlabeled foggy data, leading to insufficient model generalization ability.
To address these issues, we propose a novel training framework, \textbf{D}ecouple \textbf{D}efogging and \textbf{S}emantic learning, called $\rm D^2SL$, aiming to alleviate the adverse impact of defogging tasks on the final segmentation task. 
In this framework, we introduce a domain-consistent transfer strategy to establish a connection between defogging and segmentation tasks.
Furthermore, we design a real fog transfer strategy to improve defogging effects by fully leveraging the fog priors from real foggy images. 
Our approach enhances the semantic representations required for segmentation during the defogging learning process and maximizes the representation capability of fog invariance by effectively utilizing real fog data.
Comprehensive experiments validate the effectiveness of the proposed method.

\end{abstract}

\section{Introduction}

Semantic segmentation in foggy conditions plays a pivotal role in ensuring the safety of autonomous driving \cite{liu2022image}, garnering significant attention in recent years. 
Given the unique challenges posed by specific acquisition conditions and intricate annotation requirements \cite{li2023vblc}, \textbf{u}nsupervised \textbf{d}omain \textbf{a}daptation (UDA) methods \cite{luo2019significance,wang2020classes} have been introduced for practical implementation in this field. 
The primary goal of UDA is to transfer knowledge acquired from labeled clean data to unlabeled foggy data \cite{leefifo}, ultimately improving the model's adaptability to challenging foggy conditions.

\begin{figure}[t]
\centering
\includegraphics[width=1\linewidth]{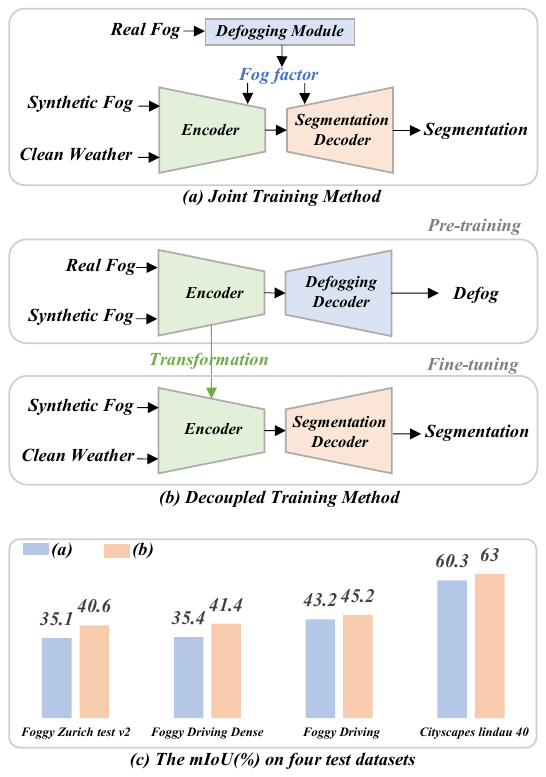}
\caption{\textbf{Impact of joint learning and decoupled learning.}  
}
\label{fig:task}
\end{figure}

Currently, state-of-the-art UDA methods in this field, e.g. FIFO\cite{leefifo}, extract fog-invariant features by aligning fog-style proxies (i.e., gram matrices) between real clean fog data and synthetic fog data, and then force the model to learn semantics and jointly defogging express, as shown in \cref{fig:task} (a).
However, this training paradigm has two drawbacks. Firstly, it couples the representation learning of semantics and defogging, complicating the segmentation task's ability to acquire precise semantic representation. 
This challenge arises from the fact that semantic segmentation demands a high-level understanding of semantics\cite{lin2017refinenet} while defogging necessitates the preservation of low-level details\cite{chen2021psd,guo2022image,liu2021synthetic,qin2020ffa}. When both are simultaneously optimized, their objectives conflict \cite{liu2022image}.
Secondly, the failure to harness real fog priors in unlabeled foggy data results in an inadequate model generalization ability. 
This limitation stems from the belief that solely learning fog representations through the synthesis of fog from clean images lacks genuine fog priors, introducing bias into the acquired fog-invariant representations.

We propose a novel training framework, \textbf{D}ecouple \textbf{D}efogging and \textbf{S}emantic learning, called $\rm D^2SL$, which learns better semantics for segmentation while keeping the defogging ability.
In $\rm D^2SL$, we introduce a Domain-Consistent Transfer (DCT) strategy to seamlessly connect defogging and segmentation tasks. As illustrated in \cref{fig:task} (b), DCT disentangles defogging and segmentation tasks by aligning features extracted by the defogging encoder with those extracted by the segmentation encoder on the corresponding clean image.
Additionally, we devise a Real Fog Transfer (RFT) strategy to optimize defogging effects by fully capitalizing on the fog priors inherent in real foggy images. 
RFT enhances the semantic features of defogging images from both synthetic and real fog datasets, bringing them into close resemblance with their respective clean images. 
We compare the effect of joint training using defogging loss and segmentation loss together in \cref{fig:task} (a) and decoupling training using these two losses separately in \cref{fig:task} (b).
\cref{fig:task} (c) shows that the mIoU of \cref{fig:task} (a) is significantly lower than that of \cref{fig:task} (b), which proves that the defogging features affect the semantic expression of fog segmentation.
Comprehensive experiments demonstrate that our method consistently delivers robust performance across various domain-adaptive tasks in foggy conditions.

In summary, our contributions are as follows:
\begin{itemize}
\item We propose a Decouple Defogging and Semantic learning ($\rm D^2SL$), which learns better semantics for segmentation while keeping the defogging ability. 

\item We introduce a Domain-Consistent Transfer (DCT) strategy to seamlessly connect defogging and segmentation task and a real fog transfer (RFT) strategy to optimize defogging effects. 

\item $\rm D^2SL$ outperforms contemporary methods and demonstrates the new state-of-the-art performance on the fog segmentation datasets.
\end{itemize}

\section{Related Work}

\subsection{Image Dehazing}
Fog images with low visibility seriously affect subjective perception and the performance of downstream tasks. Many learning-based methods for dehazing \cite{zheng2023curricular,chen2021psd,guo2022image,liu2021synthetic,qin2020ffa} have been proposed so far to restore latent clean image from foggy input. However, they are generallycomputationally complex and are not directly applicable as defogging modules before downstream tasks. 
Therefore, in order to avoid additional defogging modules, we train the model of downstream tasks to have a certain defogging ability to reduce the waste of computing power and the delay of reasoning speed.

\subsection{Unsupervised Domain Adaptation (UDA)}
UDA refers to the process of adapting a model from the source domain to an unlabeled target domain.
Most existing methods \cite{luo2019significance,tsai2018learning,wang2020classes} employ adversarial techniques to train both the segmentation network and the discriminator. 
As the discriminator's ability to maximize the difference between source and target domains increases, the segmentation model can progressively reduce this difference. 
FIFO \cite{leefifo} considers the fog condition of an image as its style and closes the gap between images with different fog conditions in neural style spaces of a segmentation model. 
All of these approaches combine the defogging ability with the segmentation for joint training, which may limit the segmentation ability of the model and wastes a lot of training resources.
Compared with them, $\rm D^2SL$ proposes to decouple the defogging task from the fog segmentation task to enhance adaptability.

\subsection{Pre-training Method}
Significant advancements have been made in generative self-supervised learning for computer vision.  
A number of studies\cite{MAE,simmim,mixmim} have focused on enhancing downstream visual tasks through the utilization of effective information from pre-text tasks.
In detail, MAE \cite{MAE} and SimMIM \cite{simmim} replace a random subset of input tokens with a special MASK symbol and aim at reconstructing original image tokens from the corrupted image. Subsequently, MixMIM \cite{mixmim} finds that using the mask symbol significantly causes training-finetuning inconsistency and replaces the masked tokens of one image with visible tokens of another image. 
However, all of these designs are based on the Vision transformers \cite{vit,swin}, which inherently have a token structure suitable for pre-text tasks.  
SparK \cite{tian2023designing} and A2MIM \cite{A2MIM} apply the idea of masked image modeling to convolutional neural networks (CNNs). But they are only designed for classification and recognition of clean images, not for domain adaption in foggy scenes.

\begin{figure*}[t]
\centering
\includegraphics[width=1\linewidth]{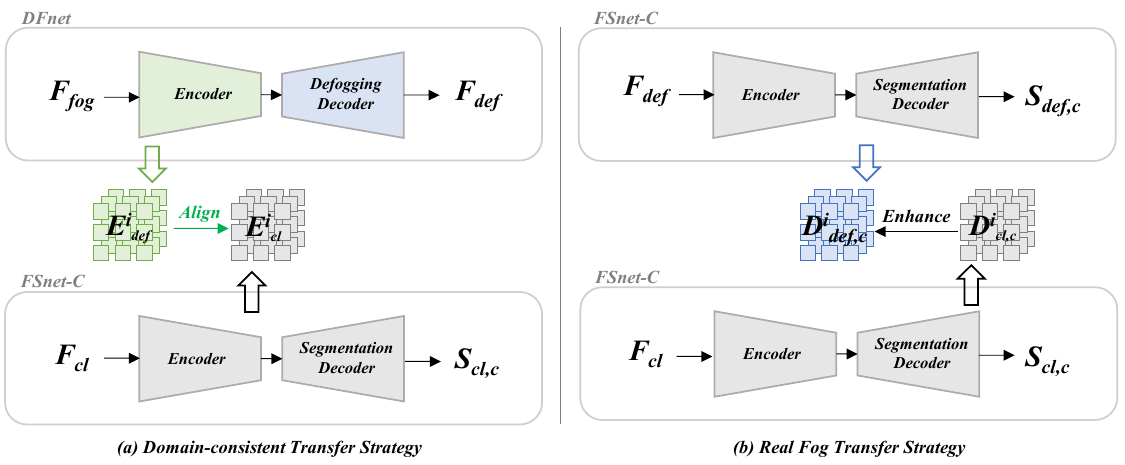} 
\caption{\textbf{An overview of $\rm D^2SL$.} 
(a) The Domain-Consistent Transfer strategy aligns features extracted by the defogging encoder with those extracted by the segmentation encoder on the corresponding clean images, thereby disentangling the defogging and segmentation tasks. (b) The Real Fog Transfer strategy enhances semantic features of defogged images from both synthetic and real fog datasets, making them highly similar to their respective clean images. By leveraging Domain-Consistent Transfer and Real Fog Transfer strategies during the pre-training phase, $\rm D^2SL$ prevents defogging features from influencing semantic expression while incorporating real fog priors.
}
\label{fig:domain2}
\end{figure*}

\section{Methodology}

\subsection{Overview}

$\rm D^2SL$ decomposes defogging learning and semantic learning into two distinct stages: defog pre-training and semantic segmentation fine-tuning. The overarching training framework for defog pre-training is depicted in Fig. 2, comprising synthetic fog pre-training and real fog pre-training, conducted sequentially in a progressively structured curriculum. As a preliminary phase, the former, employing the Domain-consistent Transfer strategy, learns a generalized defogging capability from paired synthetic-clean fog image pairs. As the primary phase, the latter introduces the Real Fog Transfer strategy, assimilating real fog data from the target domain to incorporate genuine fog priors into the pre-training, thus biasing it more toward the target domain.
Through these concerted efforts, the model acquires defogging capabilities relevant to the target domain. Driven by this capability, we introduce a semantic fine-tuning approach that facilitates direct semantic learning while preserving the defogging capability of the model.
In Sections 3.2, 3.3 and 3.4, detailed explanations of the Domain-consistent Transfer strategy, Real Fog Transfer strategy, and the fine-tuning method will be provided.


\begin{figure}[t]
\centering
\includegraphics[width=1\linewidth]{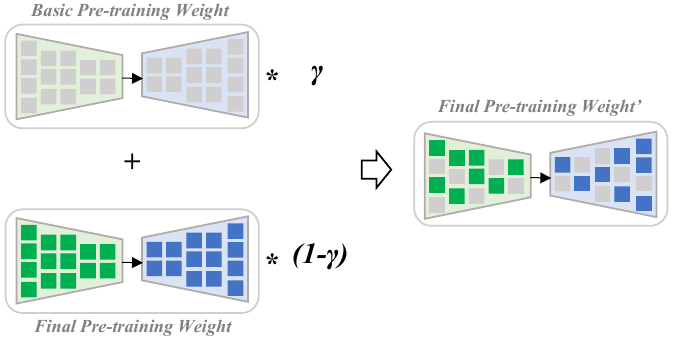}
\caption{\textbf{The training strategy of FDM.} We utilize FDM to get the fog priors inherent in real foggy images.
}
\label{fig:FDM}
\end{figure}

\subsection{Domain-Consistent Transfer Strategy}

Inspired by works such as the Masked Image Model\cite{tian2023designing,A2MIM}, we incorporate a pre-training approach to tackle domain adaptation tasks. Our objective is to equip a segmentation model with the ability to handle defogging, effectively separating defogging from semantic learning. To realize this, we employ a progressive pre-training method, initially learning a universal defogging representation from synthetically generated hazy-clean data. Subsequently, we introduce a Domain-Consistent Transfer strategy to seamlessly connect defogging and segmentation tasks, as depicted in Figure \ref{fig:domain2} (a).

Specifically, we design a loss $\mathbf{\pounds}_{DCT}$ to learn a generalized defogging capability, which can effectively assist the fog segmentation task.
We assume that $\mathbf{F}_{fog}$ is the foggy frame, $\mathbf{F}_{def}$ is the defogging frame created by the  defogging network (DFnet), and $\mathbf{F}_{cl}$ represents the clean frame paired with $\mathbf{F}_{fog}$.
$\mathbf{S}_{cl,c}$ denotes the segmentation result of the frozen segmentation network (FSnet-C) when inputting $\mathbf{F}_{cl}$.

Let $\mathbf{E}_{def}^{i}$ be the features extracted by the $\mathbf{i}^{th}$ layer of the encoder of DFnet and $\mathbf{E}_{cl}^{i}$ be the features extracted by the $\mathbf{i}^{th}$ layer of the encoder of FSnet-C.
$\mathbf{\pounds}_{DCT}$ is designed as follows:
    \begin{align}
        \label{equ:loss_seg}
        \mathbf{\pounds}_{DCT} = \sum_{i=1}^n\mathit{\pounds}\left(\mathbf{E}_{def}^{i}, \mathbf{E}_{cl}^{i}\right),
    \end{align}
where $n$ denotes that the encoder covers $n$ layers and $\mathit{\pounds}$ stands for similarity calculation.
DFnet is trained by reducing $\mathbf{\pounds}_{DCT}$, which denotes the gap between features extracted by the defogging encoder and those extracted by the segmentation encoder on the corresponding clean image.


\subsection{Real Fog Transfer Strategy}

The preceding pre-training regimen enables the model to acquire a universal defogging representation. However, this generalized representation may not yield optimal defogging effects for specific target domains due to variations in fog density and fog-inducing factors across different scenes. With this in mind, we introduce the main course, designed to integrate real fog priors into the pre-training process. This approach assists the model in learning defogging capabilities specific to the target domain. To leverage the inherent fog priors in real foggy images, we devise the Real Fog Transfer strategy, as depicted in Figure \ref{fig:domain2} (b).

We first design Fog Domain Migration (FDM) to implement synthetic fog pre-training and real fog pre-training step by step.
As shown in \cref{fig:FDM}, we adopt the synthetic paired datasets as $\mathbf{F}_{cl}$ and $\mathbf{F}_{fog}$. We train DFnet on them to get a basic pre-training weights.
Then we defog the real foggy dataset based on the basic pre-training weights to obtain the artificial defogging images.
Futher, we add them in $\mathbf{F}_{fog}$ and $\mathbf{F}_{cl}$ as the new defogging datasets.
In each iteration, we keep the base weights at a ratio $\gamma$ and pre-train again on the new defogging datasets.
In the above way, we get the final pre-trained weights.

Additionally, we design a Segmentation-Enhanced Defogging (SED) loss to ennhance the semantic features of defogging
images from both synthetic and real fog datasets, bringing them into close resemblance with their respective clean
images. 
$\mathbf{S}_{def,c}$ indicates the segmentation result of FSnet-C when inputting $\mathbf{F}_{def}$. 
Let $\mathbf{D}_{def,c}^{i}$ be the features extracted by the $\mathbf{i}^{th}$ layer of the decoder of FSnet-C and $\mathbf{D}_{cl,c}^{i}$ be the features extracted by the $\mathbf{i}^{th}$ layer.
SED $\mathbf{\pounds}_{SED}$ is given by

    \begin{align}
        \label{equ:loss_seg}
        \mathbf{\pounds}_{SED} = \sum_{i=1}^n\mathit{\pounds}\left(\mathbf{D}_{def,c}^{i}, \mathbf{D}_{cl,c}^{i}\right)+\mathit{\pounds}\left(\mathbf{S}_{def,c}, \mathbf{S}_{cl,c}\right),
    \end{align}
where $n$ denotes that the decoder covers $n$ layers.
DFnet is optimized by reducing $\mathbf{\pounds}_{SED}$, which denotes the gap between the semantic features extracted by the segmentation decoder on pairs of images.

\begin{figure}[t]
\centering
\includegraphics[width=1\linewidth]{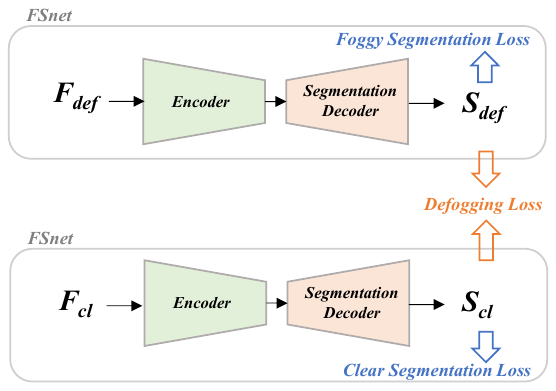}
\caption{\textbf{The fine-tuning loss.} The fine-tuning loss consists of three parts: Foggy Segmentation loss, Clean Segmentation loss, and Prediction Consistency loss. 
}
\label{fig:finetune_loss}
\end{figure}

\subsection{Fine-tuning Method}
Through these efforts, the pre-trained model has significantly improved defogging capabilities compared to its previous state. Going forward, our primary focus is on efficiently fine-tuning the pre-trained model to maintain its defogging prowess while emphasizing semantic learning.
To achieve this, we introduce a semantic fine-tuning approach that enables direct semantic learning while preserving the model's defogging capability. The fine-tuning loss $\mathbf{\pounds}{ft}$ comprises three components: Foggy Segmentation loss $\mathbf{\pounds}{fog}$, Clean Segmentation loss $\mathbf{\pounds}{cl}$, and Defogging loss $\mathbf{\pounds}{con}$, as illustrated in Figure \ref{fig:finetune_loss}.$\mathbf{\pounds}_{ft}$ can be formalized as
    \begin{align}
        \label{equ:loss_ft}
        \mathbf{\pounds}_{ft} = \mathbf{\pounds}_{fog}+\mathbf{\pounds}_{cl}+{\lambda}_{con}\mathbf{\pounds}_{con},
    \end{align}
where ${\lambda}_{con}$ is balancing hyper-parameters. For learning semantic segmentation, we apply the pixel-wise cross-entropy loss $\mathit{C}$ to individual images. 
$\mathbf{S}_{def}$ indicates the segmentation result of fog segmentation network (FSnet) when inputting $\mathbf{F}_{def}$ and $\mathbf{S}_{cl}$ indicates the segmentation result of FSnet when inputting $\mathbf{F}_{cl}$. 
To be specific, $\mathbf{\pounds}_{fog}$ and $\mathbf{\pounds}_{cl}$ are given by
    \begin{align}
        \label{equ:loss_fog}
        \mathbf{\pounds}_{fog} = \mathit{C}\left(\mathbf{S}_{def},\mathbf{Y}\right),
    \end{align}
    \begin{align}
        \label{equ:loss_cl}
        \mathbf{\pounds}_{cl} = \mathit{C}\left(\mathbf{S}_{cl},\mathbf{Y}\right),
    \end{align}
where $\mathbf{Y}$ denotes the groundtruth label.

Pairs of corresponding $\mathbf{F}_{fog}$ has the same semantic layout as $\mathbf{F}_{cl}$.
In order to ensure the defogging ability obtained by the model in the pre-training stage, we encourage the model to predict the same segmentation map while ensuring that $\mathbf{F}_{fog}$ and $\mathbf{F}_{cl}$ of the same origin. 
    \begin{align}
        \label{equ:loss_ft}
        \mathit{\pounds}_{con} = KLdiv\left(\mathbf{S}_{def},\mathbf{S}_{cl}\right),
    \end{align}
where $KLdiv$ is the Kullback–Leibler divergence.

\begin{table*}[ht]
\centering
{
\resizebox{0.9\textwidth}{!}
{%
\begin{tabular}{c|c c c|c c c|c}
    \hline
    \multicolumn{1}{c|}{Method}&
    \multicolumn{1}{c}{\makecell{Cityscapes \cite{cityscapes} \\(Clear-weather)} } &
    \multicolumn{1}{c}{\makecell{SDBF\cite{FZ} \\(Synthetic)}} &
    \multicolumn{1}{c|}{\makecell{ GoPro\cite{FZ}\\ (Real fog)} }&
    \multicolumn{1}{c}{\makecell{FZ test v2\cite{FZ}\\ mIoU ($\%$)} } &  
    \multicolumn{1}{c}{\makecell{FDD\cite{FZ}\\ mIoU ($\%$)} } & 
    \multicolumn{1}{c|}{\makecell{FD\cite{FD}\\ mIoU ($\%$)}} &
    \multicolumn{1}{c}{\makecell{CL 40\cite{cityscapes} \\ mIoU ($\%$)}} \\
    
    \hline
    RefineNet-lw\cite{refinenetlw} & \checkmark & \checkmark &  & 32.8 & 32.1 & 43.9 & 59.0   \\
    \hline
    AdSegNet\cite{tsai2018learning} & \checkmark & \checkmark & \checkmark & 25.0 & 15.8 & 29.7 & -  \\
    AdvEnt \cite{vu2019advent}& \checkmark &  \checkmark & \checkmark & 39.7 & 41.7 & 46.9 & 61.7  \\
    FDA \cite{yang2020fda}& \checkmark &  \checkmark & \checkmark & 22.2 & 29.8 & 21.8 & 39.3  \\
    DANN \cite{ganin2016domain}& \checkmark &  \checkmark & \checkmark & 43.1 & 41.4 & 46.0 & 60.1  \\
    \hline
    CMAda2+${}^{fog}$ \cite{dai2020curriculum} & \checkmark &  \checkmark & \checkmark & 43.4 & 40.1 & 49.9 & -  \\
    CMAda3+${}^{fog}$ \cite{dai2020curriculum} & \checkmark &  \checkmark & \checkmark & 46.8 & 43.0 & 49.8 & 59.6  \\
    
    \hline
    FIFO \cite{leefifo} & \checkmark &  \checkmark & \checkmark & ${42.6}^{*}$ & ${41.3}^{*}$ &  ${48.9}^{*}$ & ${66.6}^{*}$  \\  
    $\rm D^2SL$ & \checkmark & \checkmark & \checkmark& ${44.2}^{*}$ & ${42.4}^{*}$  & $45.9^{*}$ & $66.3^{*}$  \\
    
    \hline
    \end{tabular}
    }
}
\caption{\textbf{Quantitative results in mean intersection over union (mIoU).} The results is based on three real foggy datasets—Foggy Zurich test v2, Foggy Driving Dense, Foggy Driving, and a clear weather dataset—Cityscapes Lindau 40. '*' denotes that We calculate the average value according to the experimental results of repeated training for 3 times. '$fog$' means the model is trained directly on labeled foggy scenes.
} 
\label{tab:Quantitative results}
\end{table*}

\begin{figure*}[ht]
\centering
\includegraphics[width=1\linewidth]{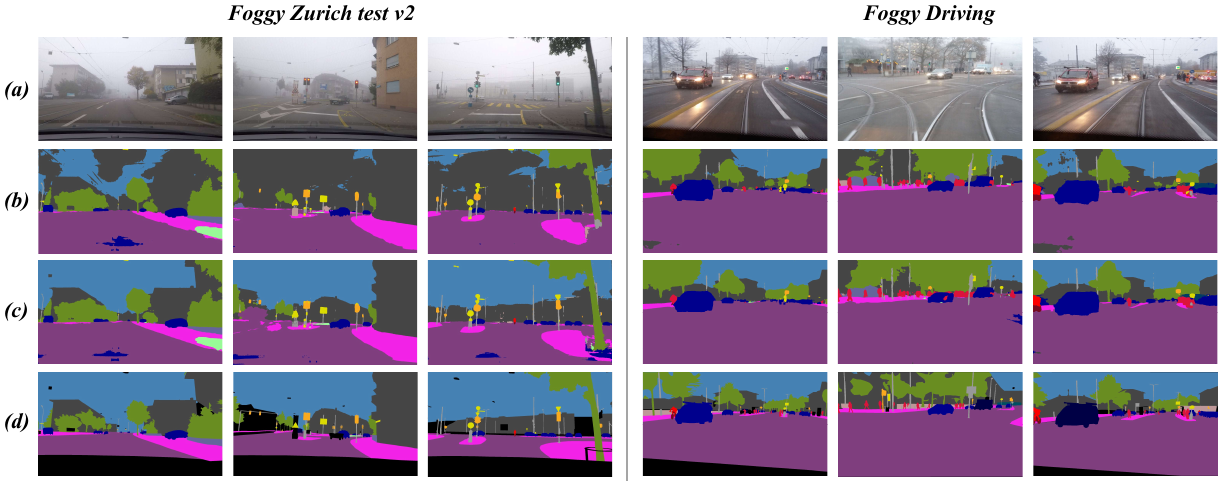} 
\caption{\textbf{Qualitative results on the real foggy datasets.}  (a) Input images. (b) Joint training. (c) $\rm D^2SL$. (d) Groundtruth.
}
\label{fig:segresult}
\end{figure*}

\section{Experiments}
\subsection{Datasets for Trainning}
We adopt the Cityscapes dataset \cite{cityscapes} as $\mathbf{F}_{cl}$, which is fully annotated for supervised learning of semantic segmentation. Meanwhile,  we utilize the Foggy Cityscapes dataset \cite{FZ} as $\mathbf{F}_{fog}$, which is constructed by simulating realistic fog effects on images of the Cityscapes dataset, thus also fully annotated. We also use the Foggy Zurich (FZ) dataset \cite{FZ} as the real foggy images during pre-training.

\subsection{Datasets for Evaluation}
Following FIFO \cite{leefifo}, we evaluate and compare $\rm D^2SL$ with previous approaches on three real-world foggy datasets: Foggy Zurich (FZ) test v2 \cite{FZ}, Foggy Driving(FD) \cite{FD}, and Foggy Driving Dense (FDD) \cite{FZ}.  These datasets consist of images depicting various levels of fog density and are fully annotated. Moreover, they share the same class set with the Cityscapes dataset described in \cite{dai2020curriculum}. Additionally, we assess the performance of both $\rm D^2SL$ and previous methods on an unseen clear weather dataset, Cityscapes Lindau (CL) 40 introduced in \cite{dai2020curriculum}, to evaluate their performance in clear weather scenes.

\subsection{Implementation Details}
$\rm D^2SL$ is implemented based on the PyTorch framework and trained with NVIDIA GeForce RTX 3090 GPUs. For pre-training, We employ Adam optimizer \cite{adam} with $\beta_{1}=0.9$, $\beta_{2}=0.99$.
We decay the learning rate from $5\times{10}^{-5}$ to $1\times{10}^{-5}$ in 50K iterations for basic pre-training and decay the learning rate from $2\times{10}^{-5}$ to $1\times{10}^{-5}$ in 20K iterations for final pre-training. The images are resized 512 × 512, and the batch size is set to 6.
Additionally, $\gamma$ is setted to 0.01. We use L1 loss to calculate the similarity of the features.

FSnet is trained by SGD with a momentum of 0.9 and the initial learning rate of $1\times{10}^{-3}$ for the encoder and $1\times{10}^{-2}$ for the decoder, both of which are decreased by polynomial decay with a power of 0.5. The input images are resized, cropped to 600 × 600, and randomly flipped horizontally. The hyper-parameter ${\lambda}_{con}$ is setted to $1\times{10}^{-4}$. The fine-tuning iterations are 60k and the batch size is 4.

We employ RefineNet-lw \cite{refinenetlw} with ResNet-101 backbone as our FSnet. As shown in \cref{fig:domain2} (a), the encoder of DFnet depends on the specific structure of FSnet. 
The decoder of DFnet consists of the Up Blocks and the Out Block. Each Up Block includes a transposed convolution layer and a convolution layer, while the Out Block is implemented with a single convolution layer.
FSnet-C is loaded with the frozen weights trained on the Cityscapes dataset \cite{cityscapes}. 
According to ResNet-101 \cite{he2016deep}, $n$ is setted to 4.

\subsection{Quantitative Analysis}

The quantitative results of $\rm D^2SL$ and previous state-of-the-art methods are presented in \cref{tab:Quantitative results}.
RefineNet-lw \cite{refinenetlw} is fine-tuned without the pre-trained weights, which we refer to as the baseline model.
Since CMAda2+ \cite{dai2020curriculum} and CMAda3+ \cite{dai2020curriculum} only focus on the fog scene and do not consider clean weather conditions, their performance is higher in the fog domain. 
To ensure a fair comparison between FIFO \cite{leefifo} and $\rm D^2SL$, we calculate the average value based on three repeated training experiments.
$\rm D^2SL$ has the best overall performance in foggy scenes with similar performance on clean weather conditions.
In particular, $\rm D^2SL$ remarkably outperforms RefineNet-lw \cite{refinenetlw}, which represents the effectiveness of pre-training.

\subsection{Qualitative Results}

\cref{fig:segresult} shows the subjective segmentation results for $\rm D^2SL$ on FZ test v2 \cite{FZ} and FD \cite{FD}.
Compared with Joint Training in \cref{fig:task} (b), $\rm D^2SL$ exhibits superior segmentation performance, which demonstrates the effectiveness of pre-training from a visualization perspective.
Especially in scenarios with high fog concentration,  $\rm D^2SL$ demonstrates enhanced semantic understanding of objects such as trees, pedestrians, and sky.

\begin{figure*}[ht]
\centering
\includegraphics[width=1\linewidth]{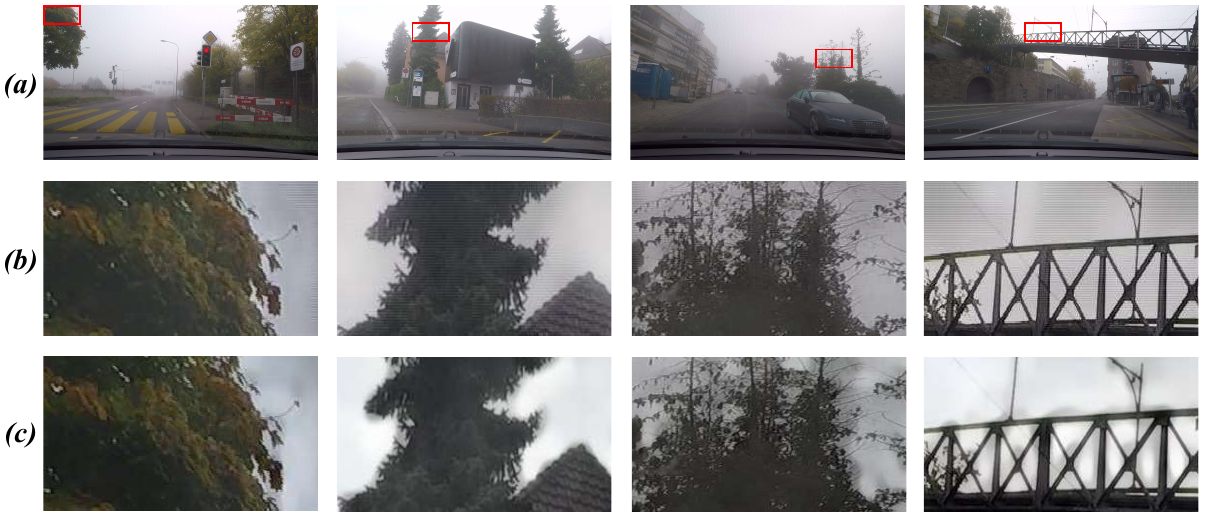} 
\caption{\textbf{Qualitative results on Foggy Zurich (FZ).}  (a) Foggy images. (b) Defogging images by the model trained with DCT and RFT. (c) Defogging images by the model trained with $\mathbf{\pounds}_{1}$.
}
\label{fig:defog}
\end{figure*}

\section{Ablation Study}
To better understand $\rm D^2SL$, we remove each critical component and assess the performance on three real foggy datasets—FZ test v2, FDD, FD, and a clear weather dataset—CL 40. $w/o$ is for not using the method. $\checkmark$ indicates that the current experiment incorporates the corresponding methods.

\textbf{Importance of components.}
In this section, we perform several ablation studies to validate the effectiveness of $\rm D^2SL$ and the necessity of every proposed method as shown in \cref{tab:ablation study of components}. 
(i) refers to RefineNet-lw \cite{refinenetlw} as presented in \cref{tab:Quantitative results}.
(ii) represents the joint training depicted in \cref{fig:task} (a), which using $\mathbf{\pounds}_{DCT}$ and $\mathbf{\pounds}_{SED}$ together. 
(iii) denotes decoupling training illustrated in \cref{fig:task} (b), where these two losses are used separately.
Notably,  both (ii) and (iii) demonstrate that the results of decoupling training is obviously better than that of joint training.
Furthermore, after employing FDM with real foggy images, objective segmentation indicators are further improved.

\begin{table}[ht]
\centering
{
\resizebox{0.9\columnwidth}{!}
{%
\begin{tabular}{c|  c c c | c  c c | c  }
    \hline
    \multicolumn{1}{c|}{Method} &
    \multicolumn{1}{c}{$\mathbf{\pounds}_{DCT}$ } &
    \multicolumn{1}{c}{$\mathbf{\pounds}_{SED}$} &
    \multicolumn{1}{c|}{FDM} &
    \multicolumn{1}{c}{FZ test v2  } &  
    \multicolumn{1}{c}{FDD } & 
    \multicolumn{1}{c|}{FD} &
    \multicolumn{1}{c}{CL 40}  \\
    \hline
    (i) &  &  &  & 32.8 & 32.1 & 43.9 & 59.0   \\
    (ii) &  \checkmark & \checkmark & &35.1 & 35.4 & 43.2  & 60.3 \\
    (iii)& \checkmark &\checkmark & & 40.6 & 41.4 & 45.2 & 63.0\\
    \hline
    $\rm D^2SL$ &  \checkmark& \checkmark & \checkmark & 44.2  &  42.4  & 45.9 & 66.3 \\
    \hline
    
    \end{tabular}
    }
}
\caption{\textbf{Impacts of the key components.} Ablation studies of each component are conducted to understand $\rm D^2SL$ better. 
}
\label{tab:ablation study of components}
\end{table}

\textbf{Different pre-training loss.}
In this part, we exhibit extensive experiments to understand how $\mathbf{\pounds}_{DCT}$ and $\mathbf{\pounds}_{SED}$ affect the final performance comprehensively in \cref{tab:ablation study of pre-training loss}.
(i) is RefineNet-lw \cite{refinenetlw} in \cref{tab:Quantitative results} without any pre-training.
Since (ii) does not impose defogging constraints on the final results of pre-training, it has no obvious effect.
Notably, (iii) demonstrates a significant improvement.
Additionally, (iv) represents the utilization of L1 loss $\mathbf{\pounds}_{1}$ during pre-training, which proves that there is the inadaptability between the defogging domain and the segmentation domain. 
However, the joint use of $\mathbf{\pounds}_{DCT}$ and $\mathbf{\pounds}_{SED}$ clearly mitigates this inadaptability while facilitating seamless knowledge transfer from defogging to segmentation domains.

In \cref{fig:defog}, we present the defogging performance  of various  pre-training loss on Foggy Zurich.
We can find that compared with $\mathbf{\pounds}_{1}$, the pre-training loss $\mathbf{\pounds}_{DCT}$ and $\mathbf{\pounds}_{SED}$ exhibit significant advantages in terms of restoring fine details such as tree branches, building railings, and house roofs. 
This observation validates that $\mathbf{\pounds}_{DCT}$ and $\mathbf{\pounds}_{SED}$ can promote the effect of segmentation tasks.

\begin{table}[ht]
\centering
{
\resizebox{1\columnwidth}{!}
{%
\begin{tabular}{c|  c c c | c  c c | c  }
    \hline
    \multicolumn{1}{c|}{Method} &
    \multicolumn{1}{c}{$\mathbf{\pounds}_{DCT}$} &
    \multicolumn{1}{c}{$\mathbf{\pounds}_{SED}$} &
    \multicolumn{1}{c|}{$\mathbf{\pounds}_{1}$} &
    \multicolumn{1}{c}{FZ test v2  } &  
    \multicolumn{1}{c}{FDD } & 
    \multicolumn{1}{c|}{FD} &
    \multicolumn{1}{c}{CL 40}  \\
    \hline
    (i) &  &  &  & 32.8 & 32.1 & 43.9 & 59.0   \\
    (ii) & \checkmark  &  &  & 32.9 & 31.5 & 42.8 & 59.9\\
    (iii)&  &\checkmark   & & 35.5 &  33.9  & 43.0 & 60.8 \\
    (iv)&  &   & \checkmark & 32.1 & 32.4 & 40.3 & 58.0\\
    \hline
    $\rm D^2SL$($w/o$) FDM &  \checkmark & \checkmark& &  40.6 & 41.4 & 45.2 & 63.0 \\
    \hline
    
    \end{tabular}
    }
}
\caption{\textbf{Impact of different pre-training loss.} 
}
\label{tab:ablation study of pre-training loss}
\end{table}

\textbf{Different fine-tuning loss.}
The performances of different fine-tuning losses are investigated in \cref{tab:ablation study of fine-tuning loss}. 
We compare the results obtained by solely utilizing $\mathbf{\pounds}_{fog}$, combining $\mathbf{\pounds}_{fog}$ and $\mathbf{\pounds}_{cl}$, and employing all three losses, respectively.
(i) proves that training the model exclusively on the fog dataset yields suboptimal performance on both the real fog datasets and the clear weather dataset.
(ii) represents that training the model on both fog and clear weather datasets enhances its robustness.
Furthermore, this robustness can be further boosted through utilization of $\mathbf{\pounds}_{con}$.

\begin{table}[ht]
\centering
{
\resizebox{1\columnwidth}{!}
{%
\begin{tabular}{c|  c c c | c c c | c  }
    \hline
    \multicolumn{1}{c|}{Method} &
    \multicolumn{1}{c}{ $\mathbf{\pounds}_{fog}$} &
    \multicolumn{1}{c}{$\mathbf{\pounds}_{cl}$} &
    \multicolumn{1}{c|}{$\mathbf{\pounds}_{con}$} &
    \multicolumn{1}{c}{FZ test v2  } &  
    \multicolumn{1}{c}{FDD } & 
    \multicolumn{1}{c|}{FD} &
    \multicolumn{1}{c}{CL 40}  \\
    \hline
    (i) & \checkmark  &  &  & 35.7  & 32.1 & 40.5 & 55.7  \\
    (ii)&  \checkmark & \checkmark   &  & 38.4 & 36.8 & 42.9 & 63.1  \\
    \hline
    $\rm D^2SL$ ($w/o$ FDM) &  \checkmark& \checkmark &  \checkmark & 40.6 & 41.4 & 45.2 & 63.0   \\
    \hline
    
    \end{tabular}
    }
}
\caption{\textbf{Impact of different fine-tuning loss.}
}
\label{tab:ablation study of fine-tuning loss}
\end{table}

\textbf{Different pre-training datasets.}
As shown in \cref{tab:ablation study of pre-training datasets},(i) and (ii) sequentially indicate that using Foggy Cityscapes dataset \cite{FZ} and FZ \cite{FZ} for pre-training can both improve the segmentation performance. 
It should be noted that FZ \cite{FZ} only consists of real fog images, and its corresponding defogging images are generated using the basic pre-trained weights that we trained on the synthetic fog dataset (Foggy Cityscapes dataset \cite{FZ}).
However, when we train them jointly in (iii), no significant enhancement is observed. 
This means that there is still some domain inadaptability between the two datasets.
By applying FDM, $\rm D^2SL$ achieves highly competitive performance according to experimental results.

\begin{table}[ht]
\centering
{
\resizebox{1\columnwidth}{!}
{%
\begin{tabular}{c|  c c  | c  c c | c  }
    \hline
    \multicolumn{1}{c|}{Method} &
    \multicolumn{1}{c}{ \makecell{Foggy Cityscapes \\dataset \cite{FZ}} } &
    \multicolumn{1}{c|}{\makecell{ Foggy Zurich\\\cite{FZ}} } &
   
    \multicolumn{1}{c}{FZ test v2  } &  
    \multicolumn{1}{c}{FDD } & 
    \multicolumn{1}{c|}{FD} &
    \multicolumn{1}{c}{CL 40}  \\
    \hline
    (i) &   & \checkmark  & 38.9 & 38.6 & 44.9 & 65.8 \\
    (ii) & \checkmark   &  & 40.6 & 41.4 & 45.2 & 63.0 \\
    (iii)& \checkmark &\checkmark  & 41.0  & 40.8 & 45.4   & 64.3   \\
    \hline
   $\rm D^2SL$ &  \checkmark & \checkmark&  44.2 &42.4  & 45.9 &66.3 \\
    \hline
    
    \end{tabular}
    }
}
\caption{\textbf{Impact of different pre-training datasets.} 
}
\label{tab:ablation study of pre-training datasets}
\end{table}

\begin{figure}[ht]
\centering
\includegraphics[width=1\linewidth]{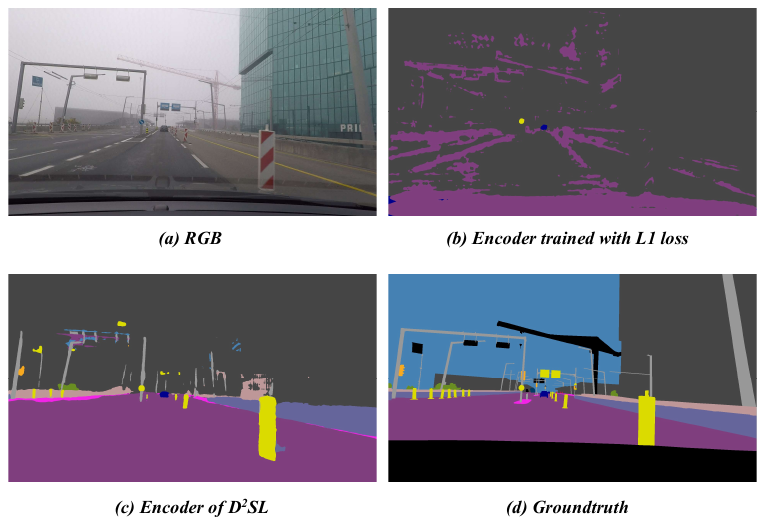}
\caption{\textbf{Segmentation results of different encoder weights.} 
}
\label{fig:intro}
\end{figure}

\textbf{The subjective segmentation results of different encoder weights.}
To demonstrate the adaptability of different pre-training to the fog segmentation task, we refrain from conducting fine-tuning training and directly merge the weights of various pre-trained encoders with the decoder weights of FSnet-C.
As shown in \cref{tab:ablation study of different encoder weights}, the encoder trained with $\mathbf{\pounds}_{1}$ does not align with the weight distribution of FSnet-C, while the pre-trained weights of $\rm D^2SL$ exhibit superior adaptability.
We evaluate these combination schemes on a fog image of FZ in \cref{fig:intro}. 
\cref{fig:intro} (c) shows that the pre-trained weights of $\rm D^2SL$ can still effectively segment roads, road signs, and some traffic lights, which shows the effect of DCT and RFT.

\begin{table}[ht]
\centering
{
\resizebox{1\columnwidth}{!}
{%
\begin{tabular}{c|  c c c | c c c | c}
    \hline
    \multicolumn{1}{c|}{Method} &
    \multicolumn{1}{c}{\makecell{Encoder\\($\rm D^2SL$)}} &
    \multicolumn{1}{c}{\makecell{Encoder\\($\mathbf{\pounds}_{1}$)}} &
    \multicolumn{1}{c|}{\makecell{Encoder\\(FSnet-C)}} &
    \multicolumn{1}{c}{FZ test v2  } &  
    \multicolumn{1}{c}{FDD } & 
    \multicolumn{1}{c|}{FD} &
    \multicolumn{1}{c}{CL 40}  \\
    \hline
    (i) &  \checkmark &  & &31.3 &  31.9 & 42.0 & 67.6 \\
    (ii)&   & \checkmark  & &2.2 & 6.9  & 4.9 & 7.5  \\
    (iii)&   &  & \checkmark &  28.5 & 35.9 & 43.6 & 63.8\\

    \hline
    
    \end{tabular}
    }
}
\caption{\textbf{Impact of different encoder weights.}
}
\label{tab:ablation study of different encoder weights}
\end{table}

\begin{figure}[ht]
\centering
\includegraphics[width=0.9\linewidth]{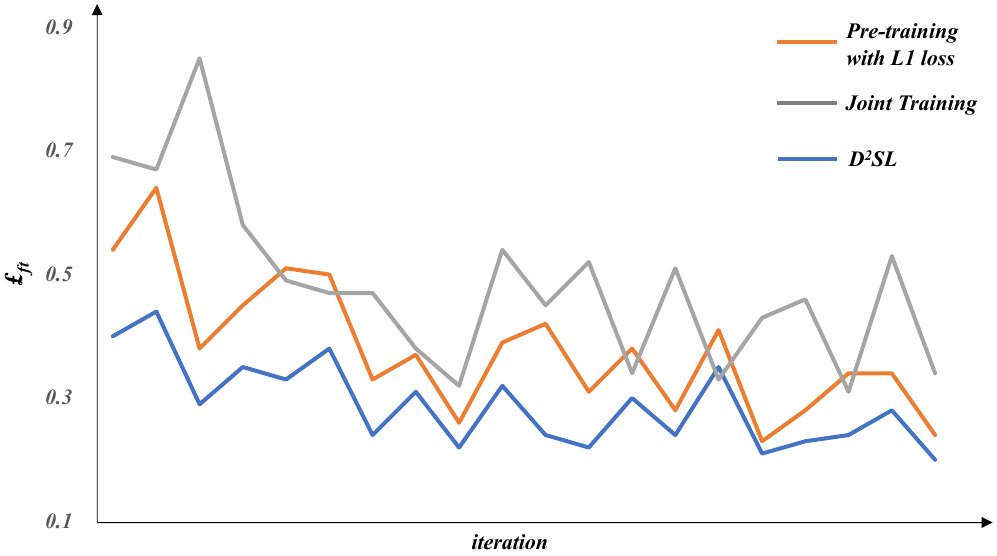}
\caption{\textbf{Fine-tuning loss of different methods.} 
}
\label{fig:loss}
\end{figure}

\textbf{Fine-tuning loss of different methods.}
The fine-tuning loss of different methods are investigated in \cref{fig:loss}. 
We compare the losses of Joint Training, Pre-training with L1 loss, and $\rm D^2SL$. 
It is evident that Joint Training results in loss oscillations due to the inconsistency of the two task optimizations.
The method of pre-training with L1 loss exhibits greater stability compared to Joint Training.
Moreover, $\rm D^2SL$ significantly accelerates the convergence of fine-tuning optimization.

\textbf{Different defogging decoders.}
To demonstrate the impact of different decoders on the final fog segmentation task, we employ two decoders with different parameter numbers in the defogging network in \cref{tab:ablation study of different defogging decoders}.
(i) is baseline without pre-training.
The decoder of (ii) adds three Resblocks \cite{he2016deep} to each layer of the decoder in section 4.3 to increase the magnitude of the decoder.
The results indicate that using a larger decoder in defogging pre-training may constrain  the semantic learning of the encoder and consequently influence fine-tuning outcomes.

\begin{table}[ht]
\centering
{
\resizebox{1\columnwidth}{!}
{%
\begin{tabular}{c|  c c  | c  c c | c  }
    \hline
    \multicolumn{1}{c|}{Method} &
    \multicolumn{1}{c}{$\mathbf{\pounds}_{DCT}$ } &
    \multicolumn{1}{c|}{$\mathbf{\pounds}_{SED}$} &
    
    \multicolumn{1}{c}{FZ test v2  } &  
    \multicolumn{1}{c}{FDD } & 
    \multicolumn{1}{c|}{FD} &
    \multicolumn{1}{c}{CL 40}  \\
    \hline
    (i) &  &  &  32.8 & 32.1 & 43.9 & 59.0   \\
    (ii) &  \checkmark & \checkmark &38.0 & 38.9 & 44.4  & 62.8 \\  
    \hline
    $\rm D^2SL$ ($w/o$ FDM)& \checkmark &\checkmark & 40.6 & 41.4 & 45.2 & 63.0\\
    \hline
    
    \end{tabular}
    }
}
\caption{\textbf{Impact of different defogging decoders.} 
}
\label{tab:ablation study of different defogging decoders}
\end{table}

\textbf{Different FSnet-C weights.}
In order to show the impact of different FSnet-C weights on the final fog segmentation task, we try to perform $\mathbf{\pounds}_{DCT}$ and $\mathbf{\pounds}_{SED}$ with different FSnet-C weights in \cref{tab:ablation study of different FSnet-C}. 
(i) is baseline without pre-training. 
For (ii), the weights of FSnet-C are the fog segmentation weights of $\rm D^2SL$ ($w/o$ FDM) instead of the weights trained on the clear weather datasets.
The fog segmentation weights hinder the alignment of encoder features in $\mathbf{\pounds}_{DCT}$ and the enhancement of decoder features in $\mathbf{\pounds}_{SED}$, resulting in performance degradation on real fog datasets.

\begin{table}[ht]
\centering
{
\resizebox{1\columnwidth}{!}
{%
\begin{tabular}{c|  c c  | c  c c | c  }
    \hline
    \multicolumn{1}{c|}{Method} &
    \multicolumn{1}{c}{$\mathbf{\pounds}_{DCT}$ } &
    \multicolumn{1}{c|}{$\mathbf{\pounds}_{SED}$} &
    
    \multicolumn{1}{c}{FZ test v2  } &  
    \multicolumn{1}{c}{FDD } & 
    \multicolumn{1}{c|}{FD} &
    \multicolumn{1}{c}{CL 40}  \\
    \hline
    (i) &  &  &  32.8 & 32.1 & 43.9 & 59.0   \\
    (ii) &  \checkmark & \checkmark &36.7 & 38.5 & 45.6  & 67.0 \\  
    \hline
    $\rm D^2SL$ ($w/o$ FDM)& \checkmark &\checkmark & 40.6 & 41.4 & 45.2 & 63.0\\
    \hline
    
    \end{tabular}
    }
}
\caption{\textbf{Impact of different FSnet-C weights.} 
}
\label{tab:ablation study of different FSnet-C}
\end{table}

\begin{figure}[ht]
\centering
\includegraphics[width=1\linewidth]{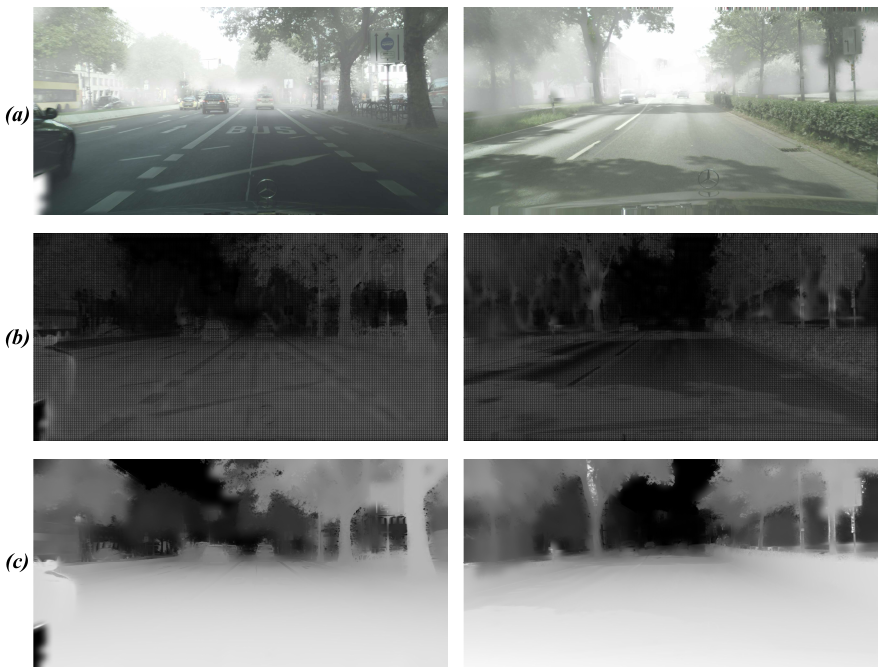}
\caption{\textbf{Depth estimation results.} (a) Foggy images. (b) Depth maps by depth estimation pre-training model. (c) Groundtruth.
}
\label{fig:depth}
\end{figure}

\textbf{Different pre-training methods.}
Since clean weather is more suitable for segmentation tasks, defogging pre-training may be helpful for fine-tuning foggy segmentation tasks, which is also demonstrated by our experimental results.
Additionally, the inherent ability of depth estimation to perform segmentation intuitively aids in improving foggy segmentation tasks. 
Therefore, we explore training a depth estimation pre-training model on the Transmittance Maps dataset \cite{FD} and subsequently fine-tuning it for foggy segmentation tasks in \cref{tab:ablation study of different pre-training methods}.
(i) is baseline without pre-training. 
Both (ii) and (iii) are pretrained on the depth estimation task. 
(ii) utilizes only $\mathbf{\pounds}_{SED}$ and (iii) utilizes both $\mathbf{\pounds}_{DCT}$ and $\mathbf{\pounds}_{SED}$.
(ii) shows that pre-training with depth estimation can improve the performance of fog segmentation task. 
Since the defogging task can be used as an intermediate step towards foggy segmentation, $\mathbf{\pounds}_{DCT}$ contributes positively to this process. 
However, the depth estimation task is not an intermediate step but rather overlaps with the fog segmentation task in the target domain, $\mathbf{\pounds}_{DCT}$ has a negative impact in this scenario. \cref{fig:depth} shows the depth estimation results of (ii).
The above experimental results fully prove the generalization of $\mathbf{\pounds}_{SED}$ across different domain adaptation.

\begin{table}[ht]
\centering
{
\resizebox{1\columnwidth}{!}
{%
\begin{tabular}{c|  c  c | c  c c | c  }
    \hline
    \multicolumn{1}{c|}{Method} &
    \multicolumn{1}{c}{$\mathbf{\pounds}_{DCT}$ } &
    \multicolumn{1}{c|}{$\mathbf{\pounds}_{SED}$} &

    \multicolumn{1}{c}{FZ test v2  } &  
    \multicolumn{1}{c}{FDD } & 
    \multicolumn{1}{c|}{FD} &
    \multicolumn{1}{c}{CL 40}  \\
    \hline
    (i) &  &  &  32.8 & 32.1 & 43.9 & 59.0   \\
    (ii) &  & \checkmark &39.9 & 34.9 & 41.6  & 63.2 \\  
    (iii)& \checkmark &\checkmark &31.3 & 32.5 & 37.8 & 58.9\\
    \hline
    $\rm D^2SL$ ($w/o$ FDM)& \checkmark &\checkmark& 40.6 & 41.4 & 45.2 & 63.0 \\
    \hline
    
    \end{tabular}
    }
}
\caption{\textbf{Impact of different pre-training methods.} 
}
\label{tab:ablation study of different pre-training methods}
\end{table}

\section{Conclusion}
We propose a novel training framework $\rm D^2SL$, aiming to alleviate the adverse impact of defogging tasks on the final segmentation task. 
In this framework, we introduce a domain-consistent transfer strategy to establish a connection between defogging and segmentation tasks.
Furthermore, we design a real fog transfer strategy to improve defogging effects by fully leveraging the fog priors from real foggy images. 
Our approach enhances the semantic representations required for segmentation during the defogging learning process and maximizes the representation capability of fog invariance by effectively utilizing real fog data.
Comprehensive experiments validate the effectiveness of the proposed method.

{\small
\bibliographystyle{ieee_fullname}
\bibliography{main}

\begin{thebibliography}{10}
\providecommand{\url}[1]{#1}
\csname url@samestyle\endcsname
\providecommand{\newblock}{\relax}
\providecommand{\bibinfo}[2]{#2}
\providecommand{\BIBentrySTDinterwordspacing}{\spaceskip=0pt\relax}
\providecommand{\BIBentryALTinterwordstretchfactor}{4}
\providecommand{\BIBentryALTinterwordspacing}{\spaceskip=\fontdimen2\font plus
\BIBentryALTinterwordstretchfactor\fontdimen3\font minus \fontdimen4\font\relax}
\providecommand{\BIBforeignlanguage}[2]{{%
\expandafter\ifx\csname l@#1\endcsname\relax
\typeout{** WARNING: IEEEtran.bst: No hyphenation pattern has been}%
\typeout{** loaded for the language `#1'. Using the pattern for}%
\typeout{** the default language instead.}%
\else
\language=\csname l@#1\endcsname
\fi
#2}}
\providecommand{\BIBdecl}{\relax}
\BIBdecl

\bibitem{Authors14}
F.~LastName, ``The frobnicatable foo filter,'' 2014, face and Gesture submission ID 324. Supplied as supplemental material {\tt fg324.pdf}.

\bibitem{Authors14b}
------, ``Frobnication tutorial,'' 2014, supplied as supplemental material {\tt tr.pdf}.

\bibitem{Alpher02}
F.~Alpher, ``Frobnication,'' \emph{IEEE TPAMI}, vol.~12, no.~1, pp. 234--778, 2002.

\bibitem{Alpher03}
F.~Alpher and F.~Fotheringham-Smythe, ``Frobnication revisited,'' \emph{Journal of Foo}, vol.~13, no.~1, pp. 234--778, 2003.

\bibitem{Alpher04}
F.~Alpher, F.~Fotheringham-Smythe, and F.~Gamow, ``Can a machine frobnicate?'' \emph{Journal of Foo}, vol.~14, no.~1, pp. 234--778, 2004.

\bibitem{Alpher05}
F.~Alpher and F.~Gamow, ``Can a computer frobnicate?'' in \emph{CVPR}, 2005, pp. 234--778.

\bibitem{zheng2023curricular}
Y.~Zheng, J.~Zhan, S.~He, J.~Dong, and Y.~Du, ``Curricular contrastive regularization for physics-aware single image dehazing,'' in \emph{Proceedings of the IEEE/CVF Conference on Computer Vision and Pattern Recognition}, 2023, pp. 5785--5794.

\bibitem{qin2020ffa}
X.~Qin, Z.~Wang, Y.~Bai, X.~Xie, and H.~Jia, ``Ffa-net: Feature fusion attention network for single image dehazing,'' in \emph{Proceedings of the AAAI conference on artificial intelligence}, vol.~34, no.~07, 2020, pp. 11\,908--11\,915.

\bibitem{guo2022image}
C.-L. Guo, Q.~Yan, S.~Anwar, R.~Cong, W.~Ren, and C.~Li, ``Image dehazing transformer with transmission-aware 3d position embedding,'' in \emph{Proceedings of the IEEE/CVF Conference on Computer Vision and Pattern Recognition}, 2022, pp. 5812--5820.

\bibitem{liu2021synthetic}
Y.~Liu, L.~Zhu, S.~Pei, H.~Fu, J.~Qin, Q.~Zhang, L.~Wan, and W.~Feng, ``From synthetic to real: Image dehazing collaborating with unlabeled real data,'' in \emph{Proceedings of the 29th ACM international conference on multimedia}, 2021, pp. 50--58.

\bibitem{chen2021psd}
Z.~Chen, Y.~Wang, Y.~Yang, and D.~Liu, ``Psd: Principled synthetic-to-real dehazing guided by physical priors,'' in \emph{Proceedings of the IEEE/CVF conference on computer vision and pattern recognition}, 2021, pp. 7180--7189.

\bibitem{A2MIM}
S.~Li, D.~Wu, F.~Wu, Z.~Zang, K.~Wang, L.~Shang, B.~Sun, H.~Li, S.~Li \emph{et~al.}, ``Architecture-agnostic masked image modeling--from vit back to cnn,'' \emph{arXiv preprint arXiv:2205.13943}, 2022.

\bibitem{MAE}
K.~He, X.~Chen, S.~Xie, Y.~Li, P.~Doll{\'a}r, and R.~Girshick, ``Masked autoencoders are scalable vision learners,'' in \emph{Proceedings of the IEEE/CVF Conference on Computer Vision and Pattern Recognition}, 2022, pp. 16\,000--16\,009.

\bibitem{simmim}
Z.~Xie, Z.~Zhang, Y.~Cao, Y.~Lin, J.~Bao, Z.~Yao, Q.~Dai, and H.~Hu, ``Simmim: A simple framework for masked image modeling,'' in \emph{Proceedings of the IEEE/CVF Conference on Computer Vision and Pattern Recognition}, 2022, pp. 9653--9663.

\bibitem{mixmim}
J.~Liu, X.~Huang, Y.~Liu, and H.~Li, ``Mixmim: Mixed and masked image modeling for efficient visual representation learning,'' \emph{arXiv preprint arXiv:2205.13137}, 2022.

\bibitem{vit}
A.~Dosovitskiy, L.~Beyer, A.~Kolesnikov, D.~Weissenborn, X.~Zhai, T.~Unterthiner, M.~Dehghani, M.~Minderer, G.~Heigold, S.~Gelly \emph{et~al.}, ``An image is worth 16x16 words: Transformers for image recognition at scale,'' \emph{arXiv preprint arXiv:2010.11929}, 2020.

\bibitem{swin}
Z.~Liu, Y.~Lin, Y.~Cao, H.~Hu, Y.~Wei, Z.~Zhang, S.~Lin, and B.~Guo, ``Swin transformer: Hierarchical vision transformer using shifted windows,'' in \emph{Proceedings of the IEEE/CVF International Conference on Computer Vision}, 2021, pp. 10\,012--10\,022.

\bibitem{tian2023designing}
K.~Tian, Y.~Jiang, Q.~Diao, C.~Lin, L.~Wang, and Z.~Yuan, ``Designing bert for convolutional networks: Sparse and hierarchical masked modeling,'' \emph{arXiv preprint arXiv:2301.03580}, 2023.

\bibitem{luo2019significance}
Y.~Luo, P.~Liu, T.~Guan, J.~Yu, and Y.~Yang, ``Significance-aware information bottleneck for domain adaptive semantic segmentation,'' in \emph{Proceedings of the IEEE/CVF International Conference on Computer Vision}, 2019, pp. 6778--6787.

\bibitem{tsai2018learning}
Y.-H. Tsai, W.-C. Hung, S.~Schulter, K.~Sohn, M.-H. Yang, and M.~Chandraker, ``Learning to adapt structured output space for semantic segmentation,'' in \emph{Proceedings of the IEEE conference on computer vision and pattern recognition}, 2018, pp. 7472--7481.

\bibitem{wang2020classes}
H.~Wang, T.~Shen, W.~Zhang, L.-Y. Duan, and T.~Mei, ``Classes matter: A fine-grained adversarial approach to cross-domain semantic segmentation,'' in \emph{European conference on computer vision}.\hskip 1em plus 0.5em minus 0.4em\relax Springer, 2020, pp. 642--659.

\bibitem{leefifo}
S.~Lee, T.~Son, and S.~Kwak, ``Fifo: Learning fog-invariant features for foggy scene segmentation. in 2022 ieee,'' in \emph{CVF Conference on Computer Vision and Pattern Recognition (CVPR)}, 2022, pp. 18\,889--18\,899.

\bibitem{cityscapes}
M.~Cordts, M.~Omran, S.~Ramos, T.~Rehfeld, M.~Enzweiler, R.~Benenson, U.~Franke, S.~Roth, and B.~Schiele, ``The cityscapes dataset for semantic urban scene understanding,'' in \emph{Proceedings of the IEEE conference on computer vision and pattern recognition}, 2016, pp. 3213--3223.

\bibitem{refinenetlw}
N.~Vladimir, S.~Chunhua, and R.~Ian, ``Light-weight refinenet for real-time semantic segmentation,'' in \emph{British Machine Vision Conference 2018, BMVC 2018}, 2018, p. 125.

\bibitem{lin2017refinenet}
G.~Lin, A.~Milan, C.~Shen, and I.~Reid, ``Refinenet: Multi-path refinement networks for high-resolution semantic segmentation,'' in \emph{Proceedings of the IEEE conference on computer vision and pattern recognition}, 2017, pp. 1925--1934.

\bibitem{FZ}
C.~Sakaridis, D.~Dai, S.~Hecker, and L.~Van~Gool, ``Model adaptation with synthetic and real data for semantic dense foggy scene understanding,'' in \emph{Proceedings of the european conference on computer vision (ECCV)}, 2018, pp. 687--704.

\bibitem{FD}
C.~Sakaridis, D.~Dai, and L.~Van~Gool, ``Semantic foggy scene understanding with synthetic data,'' \emph{International Journal of Computer Vision}, vol. 126, pp. 973--992, 2018.

\bibitem{dai2020curriculum}
D.~Dai, C.~Sakaridis, S.~Hecker, and L.~Van~Gool, ``Curriculum model adaptation with synthetic and real data for semantic foggy scene understanding,'' \emph{International Journal of Computer Vision}, vol. 128, pp. 1182--1204, 2020.

\bibitem{adam}
D.~P. Kingma and J.~Ba, ``Adam: A method for stochastic optimization,'' \emph{arXiv preprint arXiv:1412.6980}, 2014.

\bibitem{he2016deep}
K.~He, X.~Zhang, S.~Ren, and J.~Sun, ``Deep residual learning for image recognition,'' in \emph{Proceedings of the IEEE conference on computer vision and pattern recognition}, 2016, pp. 770--778.

\bibitem{vu2019advent}
T.-H. Vu, H.~Jain, M.~Bucher, M.~Cord, and P.~P{\'e}rez, ``Advent: Adversarial entropy minimization for domain adaptation in semantic segmentation,'' in \emph{Proceedings of the IEEE/CVF conference on computer vision and pattern recognition}, 2019, pp. 2517--2526.

\bibitem{yang2020fda}
Y.~Yang and S.~Soatto, ``Fda: Fourier domain adaptation for semantic segmentation,'' in \emph{Proceedings of the IEEE/CVF conference on computer vision and pattern recognition}, 2020, pp. 4085--4095.

\bibitem{ganin2016domain}
Y.~Ganin, E.~Ustinova, H.~Ajakan, P.~Germain, H.~Larochelle, F.~Laviolette, M.~Marchand, and V.~Lempitsky, ``Domain-adversarial training of neural networks,'' \emph{The journal of machine learning research}, vol.~17, no.~1, pp. 2096--2030, 2016.

\bibitem{bijelic2020seeing}
M.~Bijelic, T.~Gruber, F.~Mannan, F.~Kraus, W.~Ritter, K.~Dietmayer, and F.~Heide, ``Seeing through fog without seeing fog: Deep multimodal sensor fusion in unseen adverse weather,'' in \emph{Proceedings of the IEEE/CVF Conference on Computer Vision and Pattern Recognition}, 2020, pp. 11\,682--11\,692.

\bibitem{choi2021robustnet}
S.~Choi, S.~Jung, H.~Yun, J.~T. Kim, S.~Kim, and J.~Choo, ``Robustnet: Improving domain generalization in urban-scene segmentation via instance selective whitening,'' in \emph{Proceedings of the IEEE/CVF Conference on Computer Vision and Pattern Recognition}, 2021, pp. 11\,580--11\,590.

\bibitem{son2020urie}
T.~Son, J.~Kang, N.~Kim, S.~Cho, and S.~Kwak, ``Urie: Universal image enhancement for visual recognition in the wild,'' in \emph{Computer Vision--ECCV 2020: 16th European Conference, Glasgow, UK, August 23--28, 2020, Proceedings, Part IX 16}.\hskip 1em plus 0.5em minus 0.4em\relax Springer, 2020, pp. 749--765.

\bibitem{sakaridis2018model}
C.~Sakaridis, D.~Dai, S.~Hecker, and L.~Van~Gool, ``Model adaptation with synthetic and real data for semantic dense foggy scene understanding,'' in \emph{Proceedings of the european conference on computer vision (ECCV)}, 2018, pp. 687--704.

\bibitem{sakaridis2018semantic}
C.~Sakaridis, D.~Dai, and L.~Van~Gool, ``Semantic foggy scene understanding with synthetic data,'' \emph{International Journal of Computer Vision}, vol. 126, pp. 973--992, 2018.

\bibitem{wang2022continual}
Q.~Wang, O.~Fink, L.~Van~Gool, and D.~Dai, ``Continual test-time domain adaptation,'' in \emph{Proceedings of the IEEE/CVF Conference on Computer Vision and Pattern Recognition}, 2022, pp. 7201--7211.

\bibitem{liu2022image}
W.~Liu, G.~Ren, R.~Yu, S.~Guo, J.~Zhu, and L.~Zhang, ``Image-adaptive yolo for object detection in adverse weather conditions,'' in \emph{Proceedings of the AAAI Conference on Artificial Intelligence}, vol.~36, no.~2, 2022, pp. 1792--1800.

\bibitem{li2023vblc}
M.~Li, B.~Xie, S.~Li, C.~H. Liu, and X.~Cheng, ``Vblc: visibility boosting and logit-constraint learning for domain adaptive semantic segmentation under adverse conditions,'' in \emph{Proceedings of the AAAI Conference on Artificial Intelligence}, vol.~37, no.~7, 2023, pp. 8605--8613.

\end{thebibliography}
}

\end{document}